\documentclass[10pt,twocolumn,letterpaper]{article}

\usepackage[pagenumbers]{wacv} 

\usepackage{graphicx}
\usepackage{amsmath}
\usepackage{amssymb}
\usepackage{booktabs}
\usepackage{xcolor}

\usepackage{colortbl}
\usepackage{multirow}
\usepackage{float}

\usepackage[pagebackref,breaklinks,colorlinks]{hyperref}
\usepackage[capitalize]{cleveref}

\usepackage[capitalize]{cleveref}
\crefname{section}{Sec.}{Secs.}
\Crefname{section}{Section}{Sections}
\Crefname{table}{Table}{Tables}
\crefname{table}{Tab.}{Tabs.}

\def\wacvPaperID{2440} 
\def\confName{WACV}
\def\confYear{2025}

\begin{document}

\title{Covariance-based Space Regularization for Few-shot Class Incremental Learning}

\author{Yijie Hu$^{1}$, Guanyu Yang$^{1}$, Zhaorui Tan$^{1}$,  Xiaowei Huang$^{2}$, Kaizhu Huang$^{3}$,Qiu-Feng Wang$^{1}$\\
Xi'an Jiaotong-Liverpool University$^{1}$, Liverpool University$^{2}$, Duke Kunshan University$^{3}$\\
{\tt\small \{Yijie.Hu20, Zhaorui.Tan21\}@student.xjtlu.edu.cn, \{Guanyu.Yang02, Qiufeng.Wang\}@xjtlu.edu.cn} \\
{\tt\small xiaowei.huang@liverpool.ac.uk, kaizhu.huang@dukekunshan.edu.cn}
 }
\maketitle

\begin{abstract}
Few-shot Class Incremental Learning (FSCIL) presents a challenging yet realistic scenario, which requires the model to continually learn new classes with limited labeled data (i.e., incremental sessions) while retaining knowledge of previously learned base classes (i.e., base sessions). Due to the limited data in incremental sessions, models are prone to overfitting new classes and suffering catastrophic forgetting of base classes. 
To tackle these issues, recent advancements resort to prototype-based approaches to constrain the base class distribution and learn discriminative representations of new classes. Despite the progress, the limited data issue still induces ill-divided feature space, leading the model to confuse the new class with old classes or fail to facilitate good separation among new classes. In this paper, we aim to mitigate these issues by directly constraining the span of each class distribution from a covariance perspective. In detail, we propose a simple yet effective covariance constraint loss to force the model to learn each class distribution with the same covariance matrix. In addition, we propose a perturbation approach to perturb the few-shot training samples in the feature space, which encourages the samples to be away from the weighted distribution of other classes. 
Regarding perturbed samples as new class data, the classifier is forced to establish explicit boundaries between each new class and the existing ones.
Our approach is easy to integrate into existing FSCIL approaches to boost performance. Experiments on three benchmarks validate the effectiveness of our approach, achieving a new state-of-the-art performance of FSCIL. 

\end{abstract}

\section{Introduction}
Provided with a substantial amount of stationary data, recent advancements in deep learning have enabled neural networks to
excel in classification tasks~\cite{krizhevsky2012imagenet,he2016deep,vaswani2017attention, tan2024rethinking, tan2024interpret}. However, it is not feasible to directly deploy neural networks into the open world, where the data may emerge in a non-stationary way, such as recognizing new types of diseases or new vehicles in autonomous driving. 
Simply retraining or fine-tuning the model with new data introduces the well-known catastrophic forgetting problem~\cite{mccloskey1989catastrophic,goodfellow2013empirical}. To address this challenge, Class Incremental Learning (CIL) has been extensively researched to broaden the application scope of neural networks~\cite{zhu2021class,tian2023survey}. By simulating the scenario where disjoint new data appears in incremental sessions, CIL aims to learn new concepts without forgetting old knowledge. 

The conventional CIL setting assumes the amount of new data is usually sufficient, which may not be realistic as labeling new data can be expensive. To handle this issue, few-shot class incremental learning (FSCIL) has attracted much attention recently, where only a few training samples from new classes are available during incremental sessions~\cite{tao2020few,lesort2020continual,zhang2021few}. 
\begin{figure}
    \centering
    \includegraphics[width=\linewidth]{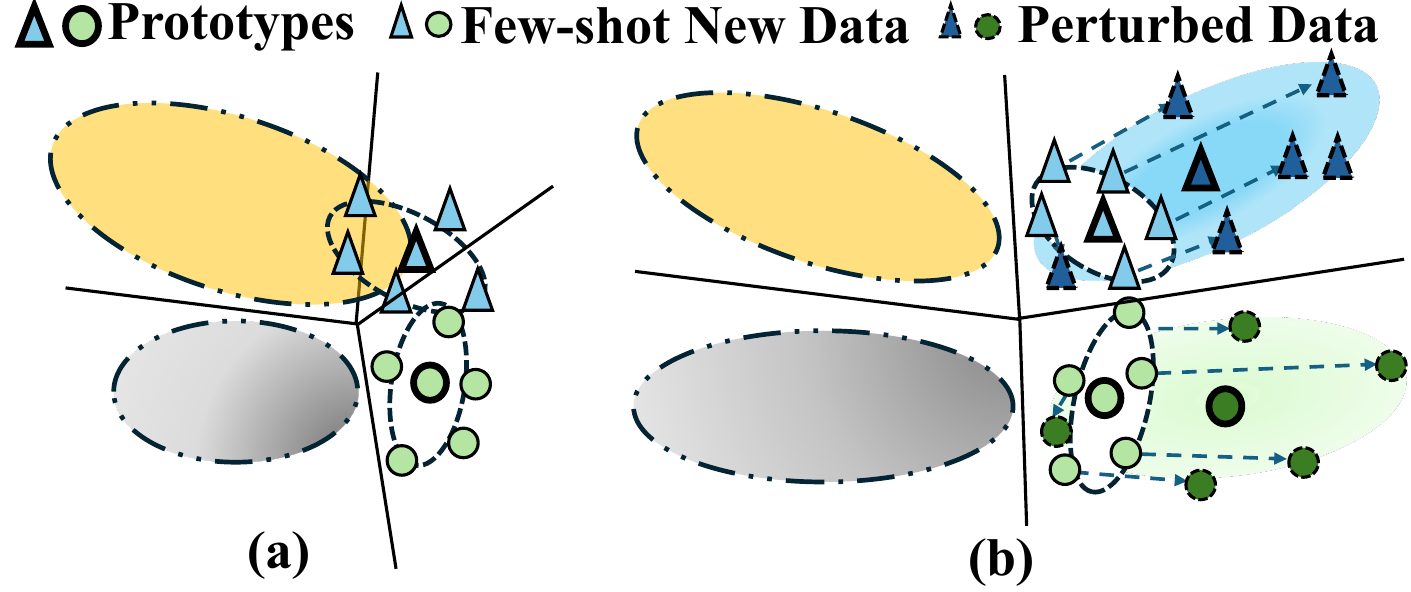}
    \caption{(a) Prototype-based models demonstrate compact representations of old classes, conserving space for new classes, though risking confusion due to mixed class distributions. 
    (b) The proposed approach aims to regularize each class distribution within a fixed span by constraining covariance and to enhance class separation through the learning of perturbed new class data.}
    \label{fig_banner}
\end{figure}
The framework of FSCIL typically involves two key stages. Initially, the model is trained on a base dataset, where all classes (referred to as base classes or old classes) contain sufficient instances. Subsequently, the model engages in incremental sessions, where it is required to learn new classes with limited samples in each session without access to previous data. After training, the model is evaluated on all previously encountered classes. Due to the scarcity of new data, the model is more vulnerable to overfit new classes and hence suffers catastrophic forgetting of old classes during incremental learning sessions~\cite{rebuffi2017icarl,li2017learning,zhu2021class, masana2022class}. 

To overcome both overfitting and catastrophic forgetting issues of FSCIL, recent wisdom resorts to prototype-based models~\cite{lesort2020continual,zhu2021self,hersche2022constrained,yang2023neural}. Specifically, 
prototype-based models~\cite{snell2017prototypical,yang2020convolutional} replace the linear classifier with learnable prototypes, aiming to learn the most representative point (e.g., center point) of each class. Recent FSCIL works involve two stages, i.e., base session learning and incremental session learning. In the base learning stage, prototype-based models are utilized to learn compact representations of base classes~\cite{peng2022few,hersche2022constrained,yang2023neural}. During the incremental sessions, the feature extractor is frozen, and new prototypes are computed using the mean of each new class. The frozen feature extractor can effectively alleviate the catastrophic forgetting problem, and the prototype classifier can mitigate the overfitting problem. However, as shown in \cref{fig_banner}\textcolor{red}{(a)}, in the fixed feature space, the estimated new class prototypes may lie very close to base class distributions due to the scarcity of the new data.  Thus, class distributions can be ill-located during incremental sessions, i.e., new class distribution may lie close to old classes or mixed with other new class distributions~\cite{yang2023neural,wang2024few}. This dilemma may then lead the model to confuse the class distributions in the feature space. Besides, the computed class prototypes can be affected by data noise, exacerbating the confusion dilemma.

To resolve this dilemma, we argue that each class should occupy the same amount of feature space. Motivated by this, 
we dynamically relocate each new class during incremental sessions to prevent overlapping. Without loss of generality, we assume each class follows the Gaussian distribution, where the mean controls the location and the covariance controls the scope of each feature distribution. To ensure each class takes up the same amount of feature space, we first regularize the covariance of each class in the base session. 
However, the direct optimization of the distribution-related statistics during training is often inefficient, as it requires model inferences throughout the entire dataset.
Inspired by previous works in variational inference~\cite{kingma2013auto,kingma2014semi,zhang2019variational}, we adopt a similar approach to estimate and regularize the class distributions efficiently. Specifically, we derive an evidence lower bound for classification and distribution learning. Maximizing the lower bound equals to maximize the confidence of prediction and minimize the KL divergence between the estimated distribution and the prior distribution. We set the fixed covariance of the prior distribution to be equal to all classes, which formulates the KL divergence to a covariance constraint loss, encouraging each class distribution to have the same covariance during training. The covariance constraint loss serves as an upper bound of KL divergence, which leads to a tighter evidence lower bound.

Furthermore, to reallocate the feature space for the few-shot new classes during incremental sessions, we propose a perturbation approach to expand the distributions of few-shot new classes by generating perturbed samples for new classes, then pushing these perturbed samples away from semantically similar classes. In detail, we introduce a learnable prior distribution for each few-shot sample based on semantic similarity. We first obtain each training sample's softmax score towards other classes, then we use the weighted mean and the fixed covariance as a prior distribution.
We adopt one linear layer to estimate the mean and variance of the new class distribution. The estimated distribution is supervised by minimizing the KL divergence between the estimated distribution and the prior. Next, we multiply the predicted variance with the extracted features and add the predicted mean to create perturbed samples, as shown in \cref{fig_banner}\textcolor{red}{(b)}. The perturbed samples are treated as new training samples, learning which pushes new classes away from the overlapping distributions. In this manner, the new class distribution is expanded by assigning the fixed variance. Learning semantic-guided perturbed samples also facilitates better separation between classes.

Our approach is easy to implement and can be integrated with other approaches. We choose several recent state-of-the-art FSCIL models as baselines~\cite{zhou2022forward,song2023learning} and apply the proposed approach to these methods. Extensive experiments on three benchmarks validate the effectiveness of our approach. 
Our main contributions can be summarized as follows:
(1) We propose a covariance constraint loss (CCL) to regularize the class distributions, constraining the span of each distribution within the same range.(2) We propose a semantic guided perturbation approach to perturb the few-shot new data, aiming to learn extensive and discriminative new class distributions. 
(3) Our proposed method is easy to incorporate into current FSCIL models to boost their performance. Experimental results on the FSCIL benchmark datasets validate the effectiveness of our approach.


\section{Related Works}
Few-shot Class Incremental Learning (FSCIL) involves the techniques of both class incremental learning and few-shot learning. 

\subsection{Class Incremental Learning} 
Class Incremental Learning (CIL) aims to learn new classes from a sequence of classification tasks without access to previously encountered data. The primary objective of CIL is to effectively learn new classes while minimizing the forgetting of old classes. Recent research in CIL can be broadly categorized into three approaches. The first and most straightforward method is to retain old data or knowledge during the learning process. Recent works~\cite{rebuffi2017icarl,kang2022class,rolnick2019experience, zhu2021class,zhou2022model} propose to mitigate the forgetting issue by rehearsing and generating previously retrained class data. Another common approach involves identifying key model parameters associated with previously learned classes and dynamically updating only the remaining parameters during incremental sessions~\cite{kim2022warping, yan2021dynamically, yoon2023soft, li2017learning}. The third category focuses on addressing the bias inherent in CIL methods, which tend to favor the most recently learned classes~\cite{wu2019large, hou2019learning, castro2018end}. 

\subsection{Few-Shot Learning} 
Few-Shot Learning (FSL) aims to develop a classification model with very limited data. To generalize on few-shot classes, metric-based methods focus on learning a similarity metric that can effectively distinguish between classes with minimal examples~\cite{vinyals2016matching,snell2017prototypical,tian2020rethinking,yu2022hybrid}. 
Hallucination-based approaches utilize data augmentation techniques, such as geometric transformations, style transfer and statistical augmentations, to increase the amount of training data~\cite{wang2019implicit,yang2021bridging,chen2022imagine,guo2022adaptive}.

\subsection{Few-Shot Class Incremental Learning}
Few-Shot Class Incremental Learning (FSCIL) aims to learn new classes with limited incoming data in an incremental manner~\cite{tao2020few, zhang2021few, zhou2022few, zhou2022forward, yang2023neural}. TOPIC~\cite{tao2020few} first introduces this setting and employs a neural gas algorithm to preserve the topology in the embedding space. To address the challenge of limited data in incremental sessions, prototype learning~\cite{snell2017prototypical, yang2020convolutional, li2023robustness} has been widely adopted in FSCIL to enhance the model's generalization to new or unseen data. To mitigate the catastrophic forgetting of old classes, recent studies~\cite{zhang2021few, hersche2022constrained, yang2023neural} propose freezing the feature extraction backbone after training the base sessions and computing new class prototypes during incremental sessions. To learn more representative prototypes, Zhu et al.~\cite{zhu2021prototype} introduces a self-promoted prototype refinement mechanism to develop extensible feature representations in the base session. LDC~\cite{liu2023learnable} utilizes a recurrent calibration module to learn new prototypes from sampled data, though this can be inefficient due to its recurrent nature. Unlike previous methods, our approach boosts FSCIL by regularizing the feature space for the few-shot new classes.

\section{Methodology}
In this section, we first give the problem formulation of FSCIL in \cref{probem}. We then describe our proposed FSCIL method in detail via two sections, i.e., Covariance Constraint Loss (CCL) in~\cref{ccl} and Semantic Perturbation Learning (SPL) in~\cref{spl}, respectively. 
\subsection{Preliminaries}
\label{probem}
\subsubsection{Problem Formulation}
FSCIL aims to train a classification model with $T$ sequential sessions $\{\mathcal{D}^{0}, \mathcal{D}^{1}, \ldots,\mathcal{D}^{T}\}$, where $\mathcal{D}^{t}=\{(\boldsymbol{x}_i^{t}, y_i^{t})\}_{i=1}^{|\mathcal{D}^{t}|}$ is the training dataset at the $t$-th session. $\boldsymbol{x}_i^{t}$ is the $i$-th input sample and its label $y_i^{t} \in \mathcal{C}^{t}$. The label space $\mathcal{C}^{t}$ of dataset $\mathcal{D}^{t}$ is disjoint between different sessions, i.e., $\forall t_1 \neq t_2, \mathcal{C}^{t_1} \cap \mathcal{C}^{t_2}=\emptyset$. The first session $\mathcal{D}^{0}$ is called the base session, which usually contains a sufficient amount of training samples for each old class $c \in \mathcal{C}^{0}$. In the next incremental session $\mathcal{D}^{t}$, there are $N$ new classes with $K$ training samples (usually 1 or 5 samples) in each class, formulating a $N$ way $K$ shot problem, i.e. $|\mathcal{D}^{t}| = N \cdot K$. In the $t$-th session, previous datasets $\{\mathcal{D}^{0}, \mathcal{D}^{1}, \ldots, \mathcal{D}^{t-1}\}$ are not available, the model can only access the data in $\mathcal{D}^{t}$. After training in session $t$, the model is evaluated on all seen classes $\tilde{\mathcal{C}}^t=\mathcal{C}^0 \cup \mathcal{C}^1 \ldots \cup \mathcal{C}^t$. 

\noindent
\subsubsection{Prototype-based Model}
\label{Prototype-based Model}
Researchers~\cite{snell2017prototypical,zhu2021self,zhou2022forward,song2023learning} commonly adopt the prototype-based framework in FSCIL, where classifier weights are treated as prototypes. During the base session, given the feature extractor $f_\psi:\boldsymbol{x}\rightarrow x\in \mathbb{R}^d$ and the classifier $\boldsymbol{W}^{0}=\{\boldsymbol{w}_1^0, \boldsymbol{w}_2^0\cdots\boldsymbol{w}_{|C^0|}^0 \}$, the model is trained using the cross-entropy loss: 
\begin{equation}
    \mathcal{L}_{ce}(\boldsymbol{x}_i^{0}, y_i^{0})=\mathbb{E}_{(\boldsymbol{x}_i^{0}, y_i^{0}) \sim \mathcal{D}^{0}} \frac{1}{N} \sum_{c=1}^{|C^0|}-y_i^{0} \log p(y=c \mid {x}_i^{0}),
\label{loss_ce}
\end{equation}
where ${x}_i^{0}=f_\psi(\boldsymbol{x}_i^{0})$ denotes the extracted feature, $p(y=c \mid {x}_i^{0})$ computes the confidence score of each sample using softmax of the cosine similarities:
\begin{equation}
     p(y=c|{x}_i^{0}) =\frac{\exp \left(\cos({\boldsymbol{w}_c} ,{x}_i^{0})\right)}{\sum_{c=1}^{\mathcal{C}^0} \exp\left( \cos({\boldsymbol{w}_c} ,{x}_i^{0})\right)}.
\end{equation}
For each incremental session $t > 0$, the feature extractor $f_\psi$ is frozen, and the classifier is updated by adding new class prototypes: $\boldsymbol{W}^{t}=\{\boldsymbol{w}_1^0, \boldsymbol{w}_2^0\cdots\boldsymbol{w}_{|C^0|}^0 \}\bigcup\{\boldsymbol{w}_1^t\ldots\boldsymbol{w}_{\mathcal{C}^t}^t\}$, where each new class prototype is computed by averaging samples from its corresponding class:
\begin{equation}
    \boldsymbol{w}_c^t = \frac{1}{|\mathcal{D}^{t}|}\sum_{i=1}^{|\mathcal{D}^{t}|}{f_\psi(\boldsymbol{x}_i^t)}.
\end{equation}

\subsection{Overview of the Framework}
\begin{figure*}
        \centering
    \includegraphics[width=\linewidth]{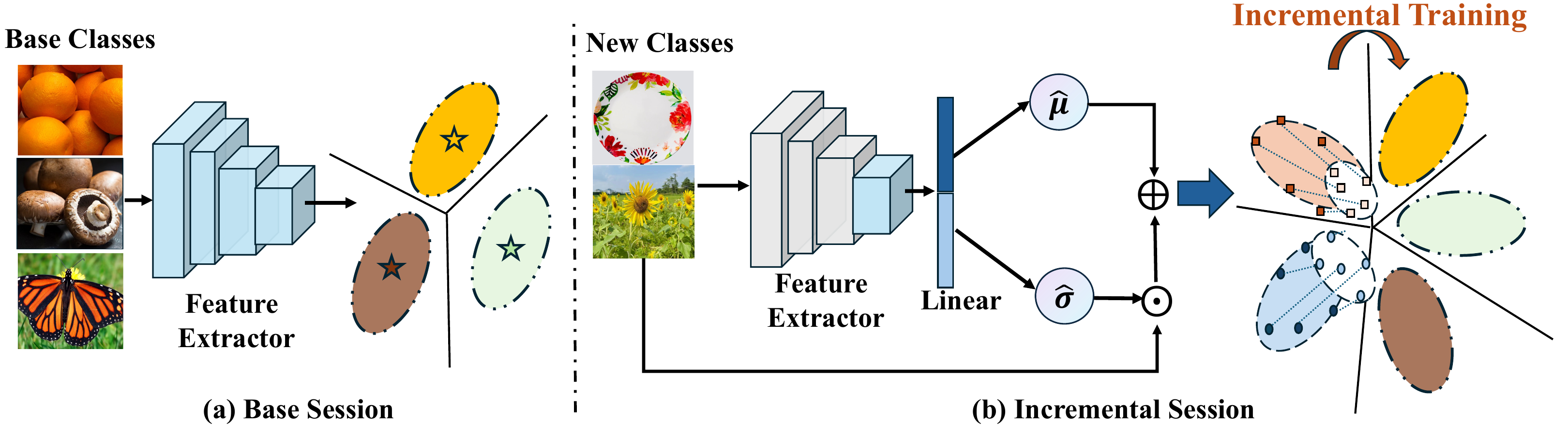}
    \caption{(a) Base session training. We deploy the covariance constraint loss to learn class distributions with a fixed span. (b) Semantic perturbation learning for the incremental stages. New data samples are perturbed by multiplying with the predicted distribution. The perturbed samples are trained along with the original data to establish the separation between classes. }
    \label{fig_appraoch}
\end{figure*}

Our framework follows the two-stage learning procedure in most of recent works~\cite{lesort2020continual,hersche2022constrained,yang2023neural}, which begins with pretraining the model using the base classes. The pretraining stage is called the base session, where we integrate the proposed covariance constraint loss (CCL) to constrain the learned representations of base classes, as shown in \cref{fig_appraoch}\textcolor{red}{(a)}. During the incremental sessions, we propose a semantic perturbation learning (SPL) approach, aiming to enlarge the separation between old and new classes by learning the perturbed new data, which is shown in \cref{fig_appraoch}\textcolor{red}{(b)}.

\subsection{Covariance Constraint Loss}
\label{ccl}
Recent wisdom~\cite{zou2022margin,chi2022metafscil,song2023learning} has demonstrated a good pretrained model benefits the learning of incoming few-shot new classes. The common attempt in recent works is to learn very compact base class representations by pushing data to the learned prototypes via fantasizing new classes or metric-based classification losses~\cite{lesort2020continual,zhou2022forward,song2023learning}. However, simply pushing the data close to the class centroid does not explicitly constrain the span of each class distribution, which may lead to representation confusion when learning new classes (\cref{fig_banner}\textcolor{red}{(a)}).
Hence, we apply a covariance constraint to each base class during training, ensuring distinct means but identical covariance across classes,
as shown in \cref{fig_appraoch}\textcolor{red}{(a)}. 

In order to constrain the covariance of each distribution, it is vital to estimate base class distributions $p(x,y)$. However, it is generally computational intractable. To solve this issue, previous works~\cite{kingma2013auto, kingma2014semi,dhuliawala2023variational} adopt the variational inference, which involves a parametric posterior function $q_{\phi}({z}|x,y)$ to approximate the true posterior by maximizing the evidence lower bound (ELBO):
\begin{equation}
\begin{aligned}
        & \log p_\theta(x,y)\\
        &= \log p_\theta(y \mid x)+\log p_\theta(z)+\log\frac{p_\theta(x \mid z)}{p_\theta(z \mid x)}\\
        &= \int q_\phi(z \mid x,y)\{\log p_\theta(y \mid x)-\log \frac{q_\phi({z}\mid{x,y})}{p_\theta(z)}+\\
        &\log \frac{q_\phi(z \mid x,y)p_\theta(x \mid z)}{p_\theta(z \mid x)}\} dz\\
        &\geq\mathbb{E}_{q_{\boldsymbol{\phi}}\left({z} \mid {x,y}\right)}\left[\log p_{\theta}\left({y} \mid {x}\right)\right]-D_{K L}\left[q_{\boldsymbol{\phi}}\left({z} \mid {x,y}\right) \| p_{\theta}({z})\right],
\end{aligned}
    \label{eq_elbo}
\end{equation}
where $\phi$ and $\theta$ are modeled by neural networks. 
The first term in \cref{eq_elbo} aims to maximize the likelihood by improving the confidence of the prediction. When optimizing the first term, the optimizing process can be written as
\begin{equation}
    \begin{aligned} 
    & \underset{q_\phi(z \mid x,y)}{\arg \max } \int_x \sum_y p(x, y) \int_z q_\phi(z \mid x,y) \log p_\theta(y \mid x) \\ = & \underset{q_\phi(z \mid x,y)}{\arg \max } \int_x p(x) \int_z q_\phi(z \mid x,y) \sum_y p(y \mid x) \log p_\theta(y \mid x).
   \end{aligned}
\end{equation}
As  $q_\phi({z}\mid{x,y})$ is a probability distribution, the integral is upper bounded by $\max\sum_y p(y \mid x) \log p_\theta(y \mid x)$, which is equivalent to minimizing the cross entropy loss in \cref{loss_ce}. The second KL divergence minimizes the divergence between the variational posterior distribution $q_{\boldsymbol{\phi}({z \mid x})}$ and the prior distribution $p_{\theta}{(z)}$. The third term actually can be calculated by $D_{K L}\left[q_{\boldsymbol{\phi}}\left({z} \mid {x,y}\right) \| p_{\theta}({z \mid x})/p_{\theta}({x \mid z})\right]$, which is a non-negative value and we eliminate this item. 
In order to explicitly contain the covariance of all distributions, the remaining problem lies in how to formulate the posterior distribution  $q_{\boldsymbol{\phi}({z \mid x})}$ and the prior distribution $p_\theta(z)$.

We first assume that $z$ satisfies multivariate Gaussian with a mean and diagonal covariance. The $p_\theta(z)$ can be formulated as $\mathcal{N}(\mu_c, \mathbf{I})$ with mean $\mathbf{\mu_c}$ as the class center and fixed eye matrix $\mathbf{I}$, respectively. As we aim to constrain the covariance, we adopt the fixed covariance as the prior.  In order to learn the latent distribution quickly from a single instance during training, we adopt linear layers 
$P_{\mu}(\cdot)$
and 
$P_{\sigma}(\cdot)$
after the feature extraction network to form a statics prediction pipeline to estimate $q_{\phi}(z|x,y)$ following previous works~\cite{kingma2013auto}
\begin{equation}
\begin{aligned}
        q_{\phi}(z|x,y)&=\mathcal{N}(z|\hat{\mu_c}, \hat{\sigma_c}^2) \\
        &=[P_{\mu}(x), P_{\sigma}(x)]_{y=c}.
\end{aligned}
\label{eq_statics_prediction}
\end{equation}
After obtaining the estimated statistical information, we can then formulate the second KL divergence term into the covariance constraint loss by
\begin{equation}
    \begin{aligned}
        &-D_{K L}\left[q_{\boldsymbol{\phi}}\left({z} \mid {x,y}\right) \| p_{\theta}({z})\right] \\
         & =\int q_\phi({z}|{x,y}) \log \frac{p_{\theta}({z})}{q_\phi({z}|{x,y})} d {z} \\
        & =\int \mathcal{N}\left({z} ; \hat{\mu}, \hat{\sigma}^2\right) \log \frac{\mathcal{N}({z} ; {\mu_c}, \mathbf{I})}{\mathcal{N}({z} ; \hat{\mu_c}, \hat{\sigma_c}^2)} d {z} \\
    &= \frac{1}{2} \sum_{c=1}^{\mathcal{C}^0}\sum_{i=1}^d \left( 1 + \log \hat{\sigma}_{c,i}^2-\hat{\sigma}_{c,i}^2 -(\hat{\mu_c}-\mu_c)^2\right)\\
    &\leq \frac{1}{2} \sum_{c=1}^{\mathcal{C}^0}\sum_{i=1}^d \left( 1 + \log \hat{\sigma}_{c,i}^2-\hat{\sigma}_{c,i}^2\right),
    \end{aligned}
    \label{eq_ccl}
\end{equation}
where $d$ represents the feature dimension. We drop the final term to formulate the covariance constraint loss as $\mathcal{L}_{ccl} =\frac{1}{2} \sum_{c=1}^{\mathcal{C}^0}\sum_{i=1}^d \left( 1 + \log \hat{\sigma}_{c,i}^2-\hat{\sigma}_{c,i}^2\right)$, which constrains the covariance of each feature distribution. \cref{eq_ccl} shows that the covariance constraint loss can be viewed as the upper bound of $-D_{KL}$. Thus, by deploying covariance constraint loss, we are able to obtain a tighter upper bound than \cref{eq_elbo}, which can be easier to optimize. The overall learning objective for the base session training is
\begin{equation}
    \mathcal{L}_{Base} = \mathcal{L}_{ce} +  \gamma\mathcal{L}_{ccl},
    \label{eq_base_loss}
\end{equation}
where $\gamma$ is a positive hyperparameter that controls the degree of constraint. 



\subsection{Semantic Perturbation Learning}
\label{spl}
In order to prevent the model from overfitting the few-shot new classes during incremental sessions, recent approaches~\cite{lesort2020continual,hersche2022constrained,song2023learning} attempt to freeze the feature extractor and learn the new class prototypes by averaging the few-shot samples. Though the overfitting problem is mitigated by fixing the previously learned feature space,  the new class distributions can only account for small parts of the feature space compared to base classes due to the scarcity of data.  Under this ill-divided space dilemma, as shown in~\cref{fig_banner}(b), new class prototypes may lie very close to base class distributions or new class prototypes, which may further lead the model to misclassify new classes. 

In this paper, we propose to address this issue from two perspectives. Firstly, we intend to expand the new class feature distributions by assigning the fixed covariance, which is the same as the base distributions and allows the new class distributions to take up more feature space. Secondly, to facilitate better separation among classes, we push the few-shot samples to be away from semantically similar distributions and then retrain the classifier to distinguish the perturbed new class samples from other classes. We achieve these two goals by forming the semantic perturbation learning framework, where we aim to learn a perturbation distribution from which any perturbations can perturb new data to be away from those close distributions but within a fixed range. The perturbation distribution can also be learned by formulating a similar variational inference in \cref{eq_elbo} but with a different prior distribution. Specifically, we form the prior distribution as the Gaussian distribution $p_{\theta}({\tilde{z}})$ as $\mathcal{N}(\tilde{\mu}, \mathbf{I})$, where $\tilde{\mu}$ is the linear combination of other class prototypes. The weight is calculated by the similarity score over classes. For each incremental session, we first initialize the classifier by computing the new class prototypes by averaging the few-shot samples. Given new data sample $\boldsymbol{x}_i^t$ with label $\boldsymbol{y}_i^t$, we compute the similarity score over other classes:
\begin{equation}
    S_{i,c} = \frac{\mathbb{I}_{c\neq{y}_i^t}\exp \left(cos({\boldsymbol{w}_c} ,{x}_i^t)\right)}{\sum_{c=1}^{|\mathcal{C}^t|-1} \exp \left(cos({\boldsymbol{w}_c} ,{x}_i^t)\right)}, 
\end{equation}
where $S \in \mathbb{R}^{K\times(|\mathcal{C}^t|-1)}$ and $\mathbb{I}(\cdot)$ is an indicator function. For each sample, we can get the prior mean by multiplying the confidence score with other class prototypes
\begin{equation}
    \tilde{\mu} = \sum_{c\neq {y}_i^t}^{|\mathcal{C}^t|-1}S_{i,c} \boldsymbol{w}_c.
\end{equation}
We also adopt a linear layer to predict the mean and covariance of each sample, the same as \cref{eq_statics_prediction}. To be noted, we do not reuse the linear layer in the base session so that this method can be directly applied to other methods in a plug-and-play manner. After predicting the statistics, the overall objective during the incremental training is:
\begin{equation}
\begin{aligned}
    \mathcal{L}_{incre} &=  \mathcal{L}_{ce}({x}_i^{t}, y_i^{t}) + \mathcal{L}_{ce}(\hat{\mu}+\hat{\sigma} \odot {x}_i^{t}, y_i^{t})\\
    &-\alpha D_{K L}\left[q_{\boldsymbol{\phi}}\left({\tilde{z}} \mid {x}_i^{t}\right) \| p_{\theta}(\tilde{z}\mid {x}_i^{t})\right],
\end{aligned}
\label{eq_spl}
\end{equation}
where $\odot$ represents the element-wise multiplication. The first and second term is the classification loss for the few-shot data and the perturbed few-shot data. The third term aims to learn the perturbation distribution by minimizing the KL divergence between the predicted feature distribution of each sample and the prior distribution. The positive hyperparameter $\alpha$ controls the strength of the perturbation. The KL term is formulated as:
\begin{equation}
\begin{aligned}
        &-D_{K L}\left[q_{\boldsymbol{\phi}}\left({\tilde{z}} \mid {x}_i^{t}\right) \| p_{\theta}(\tilde{z}\mid {x}_i^{t})\right]\\
        &= \frac{1}{2} \sum_{i=1}^d \left( 1 + \log \hat{\sigma}_i^2-\hat{\sigma}_i^2 -(\hat{\mu_i}-\tilde{\mu_i})^2 \right).
\end{aligned}
\end{equation}
\section{Experiments}
\subsection{Implementation Details}

\noindent\textbf{Datasets and Experimental Settings.} 
We conduct extensive experiments on three benchmark datasets, including MiniImageNet, CIFAR100 and CUB200. MiniImageNet and CIFAR100 contain 100 classes in total, we set the number of base classes as 60. We set 8 incremental sessions, where each session formulates a 5-way 5-shot problem. For fine-grained data CUB200 that contains 200 classes, we set the number of base classes as 100, followed by 10 incremental sessions, and each session formulates a 10-way 5-shot problem. To make the comparison fair, we use the same base and incremental class data in each dataset when conducting the experiments following previous works~\cite{agarwal2022semantics,zhou2022few, zhou2022forward,song2023learning}. All experiments are conducted using one RTX3090 card\footnote{\url{https://github.com/tambourine666/Covariance-Space-Regularize}}.

\noindent\textbf{Baseline Models.}
We adopt three baseline models, i.e., the naive CE-trained model, fantasy-based model FACT~\cite{zhou2022forward}, and state-of-the-art contrastive learning model SAVC~\cite{song2023learning}. Details of the baseline models can be found in \cref{Details_of_Baseline_Models}.

\noindent\textbf{Model Architectures.}
Following~\cite{lesort2020continual,zhou2022few,song2023learning,wang2024few}, we use ResNet-18 in experiments on CIFAR100 and MiniImageNet. We use ImageNet pre-trained ResNet-18 for the CUB200 dataset following~\cite{zhou2022few,song2023learning,wang2024few}. We follow the same experimental setting as baseline methods for fair comparisons. The dimension $d$ is set as 64, 512, 512 on CIFAR100, MiniImageNet and CUB200, respectively, which is the same as baseline models. The hyperparameter analysis is shown in \cref{abl_hyper}.  

\subsection{Benchmark Performance}
\begin{table*}[ht]
\caption{Incremental learning performance on MiniImageNet under 5-way 5-shot setup. “Avg Acc.” represents the average accuracy of all sessions. “Final Improv.” calculates the improvement of our method after learning in the final session. \textbf{Bold} represents best performance. $*$ indicates that we reproduce the results using public open-source code.}
\centering
\resizebox{\linewidth}{!}{
\begin{tabular}{@{}clccccccccccc@{}}
\toprule
\multicolumn{2}{c}{\multirow{2}{*}{\textbf{Methods}}} & \multicolumn{9}{c}{\textbf{Accuracy in each session (\%) $\uparrow$}} &  \multirow{2}{*}{\textbf{\begin{tabular}[c]{@{}c@{}}Avg\\  Acc.\end{tabular}}} & \multirow{2}{*}{\textbf{\begin{tabular}[c]{@{}c@{}}Final\\ Improv.\end{tabular}}}  \\ \cmidrule(lr){3-11}
\multicolumn{2}{c}{} & \textbf{0} & \textbf{1} & \textbf{2} & \textbf{3} & \textbf{4} & \textbf{5} & \textbf{6} & \textbf{7} & \textbf{8} &  &  \\ \midrule
 &iCaRL~\cite{rebuffi2017icarl} & 61.31 & 46.32 & 42.94 & 37.63 & 30.49 & 24.00 & 20.89 & 18.80 & 17.21 & 33.29 &+39.20  \\
&NCM~\cite{hou2019learning}& 61.31 & 47.80 & 39.30 & 31.90 & 25.70 & 21.40 & 18.70 & 17.20 & 14.17 & 30.83 & +42.24\\
 & Data-free Replay~\cite{liu2022few}& 71.84 & 67.12 & 63.21 & 59.77 & 57.01 & 53.95 & 51.55 & 49.52 & 48.21 & 58.02 &+8.20\\ 
&CEC~\cite{lesort2020continual} & 72.00 & 66.83 & 62.97 & 59.43 & 56.70 & 53.73 & 51.19 & 49.24 & 47.63 & 57.75 & +8.78\\
&MetaFSCIL~\cite{chi2022metafscil} & 72.04 & 67.94 & 63.77 & 60.29 & 57.58 & 55.16 & 52.90 & 50.79 & 49.19 & 58.85 & +7.22\\
&C-FSCIL~\cite{hersche2022constrained} & 76.40 & 71.14 & 66.46 & 63.29 & 60.42 & 57.46 & 54.78 & 53.11 & 51.41 & 61.61 &  +5.00\\
 &LIMIT~\cite{zhou2022few} & 73.81        & 72.09 & 67.87 & 63.89        & 60.70 & 57.77 & 55.67 & 53.52 & 51.23 & 61.84  &  +5.18   \\ \midrule
& CE & 75.65 & 70.45 & 66.09 & 62.16 & 58.96 & 55.92 & 53.08 & 51.05 & 49.39 & 60.31 & +7.02\\
\rowcolor{gray!20}
&\textbf{CE-Ours}& 75.77 & 70.71 & 66.53 & 63.33 & 60.42 &57.46 & 54.61& 52.48 & 50.65  & 61.33 &+5.76\\\midrule
& FACT$^{*}$~\cite{zhou2022forward}& 76.25 & 70.91 & 66.41 & 62.79 & 59.45 & 56.22 & 53.37 & 51.21 & 49.48 & 60.68&+6.93\\
\rowcolor{gray!20}
& \textbf{FACT$^{*}$-Ours}& 75.65 & 70.63 & 66.79 & 63.20 & 59.72 & 56.68 & 53.78 & 51.53 &50.01 & 60.89 & +6.40\\\midrule
& SAVC$^{*}$~\cite{song2023learning} &80.68 & 75.89 & 71.54 & 67.80 & 64.85 & 61.42 & 58.38 & 56.43 & 54.91 & 65.77&+1.50\\
\rowcolor{gray!20}& \textbf{SAVC$^{*}$-Ours}& \textbf{80.90} & \textbf{75.89} &\textbf{71.80} & \textbf{68.59} & \textbf{65.86} & \textbf{62.41} & \textbf{59.33 }&\textbf{ 57.71} & \textbf{56.41}& \textbf{66.54} &\\
\bottomrule
\end{tabular}}
\label{tab_mini}
\end{table*}

\begin{figure*}[ht]
    \centering
    \includegraphics[width=\linewidth]{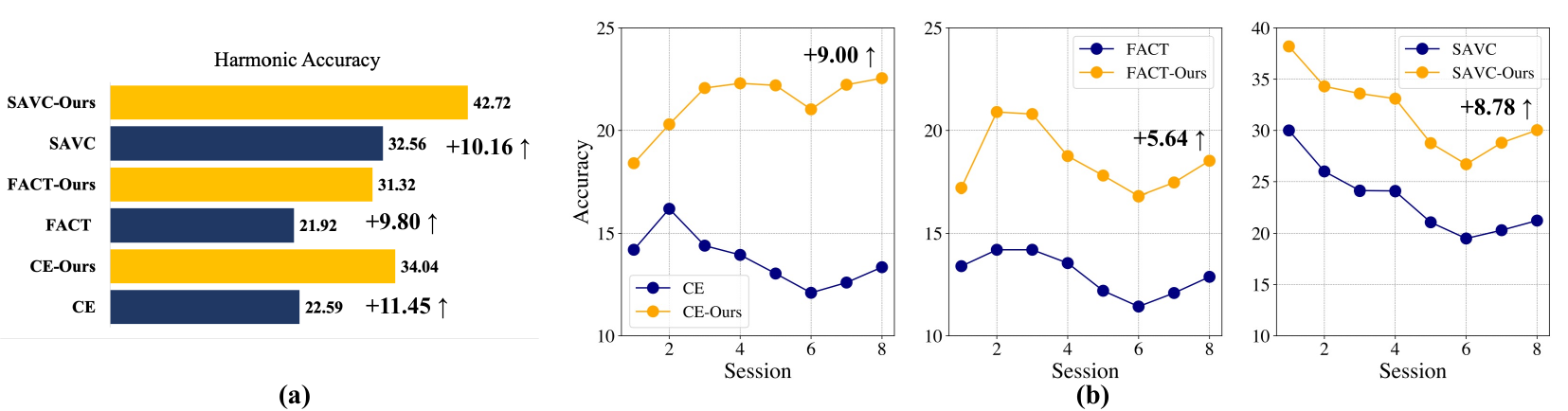}
        \caption{(a) Comparison of harmonic performance after incremental sessions with baseline models on MiniImageNet. (b) Comparison of performance of new classes with baseline model in each incremental session on MiniImageNet} 
    \label{fig_old_new_performance}
\end{figure*}
We conduct our experiments on three FSCIL benchmarks, i.e., CIFAR100,  MiniImageNet, and CUB200, and compare our approach with the baseline models and other recent FSCIL methods. We also compare our method with classical CIL methods such as ICARL~\cite{rebuffi2017icarl}, NCM~\cite{hou2019learning}, and FSCIL methods CEC~\cite{lesort2020continual}, C-FSCIL~\cite{hersche2022constrained} and meta-learning based approaches~\cite{cheng2021meta,zhou2022few}. We report the numerical results of MiniImageNet in \cref{tab_mini}, and more results on CIFAR100 (\cref{tab_cifar}) and CUB200 (\cref{tab_cub}) can be founded in \cref{more_experiments}. 
    
For baseline models, adding our proposed loss and semantic perturbation learning boosts the model performance for incremental sessions. After 8-session incremental learning, our method can improve the final model performance by at least 0.53\% on MiniImageNet and 0.56\% on CIFAR100, respectively. It is worth noticed that our method has a more profound effect on SAVC,  which learns representations by pushing the data to the class center using contrastive learning. As SAVC does not consider the span of each class distribution explicitly, constraining the class distribution using our proposed methods boosts the base class performance and obtains better performance when learning new classes. 

We further demonstrate the comparisons of harmonic performances of all classes after incremental sessions and comparisons of new class performance in each incremental session in ~\cref{fig_old_new_performance}. As shown in \cref{fig_old_new_performance}(a), compared with baseline models, our approach boosts the harmonic performance by at least 9.80\%. The CE model enjoys an improvement of 11.45\% from our approach. The significant boost in harmonic performance results from the improvement of new classes.  As illustrated in \cref{fig_old_new_performance}(b), the performance of new classes demonstrates a substantial improvement at each incremental session. Overall, our approach boosts the new class performance by at least 5.65\%. Benefiting from the covariance constraint and distribution expansion strategy, we are able to learn more separable representations of new classes, boosting the performance of new classes.

\subsection{Ablation Studies of Each Component}
\begin{table}[ht]
\caption{Ablation results of proposed components. ``Base'' represents the base model performance. ``Old'' is the performance of base classes after incremental sessions. ``New'' is the performance of all new classes. ``Avg'' is the average performance of all incremental sessions. ``PD'' is the drop rate of the base class performance after incremental sessions. ``H.'' is the harmonic accuracy of base and new classes after incremental learning.}
\resizebox{\linewidth}{!}{
\begin{tabular}{@{}c|cc|cccccc@{}}
\toprule
\multicolumn{1}{c|}{\multirow{2}{*}{\textbf{Method}}} & \textbf{\begin{tabular}[c]{@{}c@{}}Base \\ Session\end{tabular}} & \multicolumn{1}{c|}{\textbf{\begin{tabular}[c]{@{}c@{}}Incremental \\ Session\end{tabular}}} & \multicolumn{6}{c}{\textbf{CIFAR100}} \\ \cmidrule(l){2-9} 
\multicolumn{1}{c|}{} & $\mathcal{L}_{ccl}$ & \multicolumn{1}{c|}{SPL} & \textbf{Base$\uparrow$} & \textbf{Old$\uparrow$} & \textbf{New$\uparrow$} & \textbf{Avg$\uparrow$} & \textbf{PD$\downarrow$} & \textbf{H. $\uparrow$} \\ \midrule
\multicolumn{1}{c|}{CE} &  & \multicolumn{1}{c|}{} & 76.87 & 71.15 & 20.65 & 62.12 & 5.72 & 32.01 \\
CE+SPL &  &\checkmark  & 76.87 &  69.32 & 23.38 & 62.20 & 7.55 & 34.96 \\
CE$+\mathcal{L}_{ccl}$ & \checkmark &  & 78.26 & \textbf{72.67} & 21.90 & 63.58 & \textbf{5.59} & 33.66  \\
\rowcolor{gray!20}
\textbf{CE}$+\mathcal{L}_{ccl}+SPL$ & \checkmark & \checkmark  & \textbf{78.26} & 71.42 & \textbf{23.62} &\textbf{63.66}& 6.84  & \textbf{35.51} \\ \midrule
 FACT &  & & 78.38 & 71.08 & 21.73 & 62.17 & 5.40 & 32.54  \\
FACT+SPL &  & \checkmark & 78.38 & 71.07 & 21.82 & 62.20 & 7.31 & 33.40 \\
FACT+$\mathcal{L}_{ccl}$&\checkmark & &79.12 & \textbf{72.17} & 22.50 &62.80  & \textbf{4.73} & 34.31 \\
\rowcolor{gray!20}
\textbf{FACT}+$\mathcal{L}_{ccl}$+SPL& \checkmark & \checkmark & \textbf{79.12 }& 71.20 & \textbf{24.05} & \textbf{62.82} & 5.70 & \textbf{35.96} \\ \midrule
 SAVC&  &  & 78.98 & 72.25 & 21.00 & 62.81 & 6.73 & 32.56 \\ 
SAVC+SPL  &  & \checkmark  & 78.98   & 72.20 & 21.25 &62.88& 6.78 & 32.83 \\ 
SAVC+$\mathcal{L}_{ccl}$ &\checkmark  &  & 79.00 & 72.88 & 22.38& 63.35& 6.12  &34.24  \\
\rowcolor{gray!20}
\textbf{SAVC}+$\mathcal{L}_{ccl}+SPL$ & \checkmark & \checkmark & \textbf{79.00} & \textbf{72.90} & \textbf{23.50} & \textbf{63.48} & \textbf{6.10} & \textbf{35.51}\\\bottomrule
\end{tabular}
\label{tab_abl}
}
\end{table}
We conduct the ablation experiments of the proposed CCL and SPL on the CIFAR100 dataset, as shown in~\cref{tab_abl}. Following previous works~\cite{zhou2022forward,song2023learning,wang2024few}, we verify the effectiveness of our approach by six metrics, i.e., the performance of the base model, the performance on the base classes after incremental sessions, the performance of all new classes, the average performance of all incremental sessions, the performance drop of the base classes ($PD=Accuracy_{base}-Accuracy_{old}$), and the harmonic accuracy of base and new classes after the incremental learning. When we do not deploy the SPL for incremental sessions, the model adopts the prototype-based incremental learning approach mentioned in~\cref{Prototype-based Model}.

As shown in \cref{tab_abl}, by deploying our proposed covariance constraint loss $L_{ccl}$, our model is able to achieve higher base class performance compared to the baseline model. By constraining the base class distributions, the drop in PD rate and the gain of new class performance on three baseline models validate that the covariance constraint loss allows the model to learn the new classes better while obtaining less forgetting of old classes. 

We also conduct experiments of SPL on both baseline models and  $\mathcal{L}_{ccl}$ trained models. The results demonstrate that directly deploying SPL can boost the performance of new classes and harmonic accuracy by at least 0.25\% and 0.27\%, respectively. It is worth noticing that simply using SPL on the baseline models leads to a higher PD. As we expand the distribution of new classes during incremental sessions without constraining the base class distributions, it may lead the model to confuse the old classes with new classes. If SPL is deployed on the $\mathcal{L}_{ccl}$ trained model, the model is able to achieve the highest performance on new classes and the highest harmonic accuracy, validating the effectiveness of the combination of proposed components.  We also conduct experiments on comparing SPL with other incremental update methods in \cref{abl_incremental}.

\subsection{T-SNE Visualizations of Proposed Approaches}
\label{tsne}
\begin{figure}[ht]
    \centering
    \includegraphics[width=0.8\linewidth]{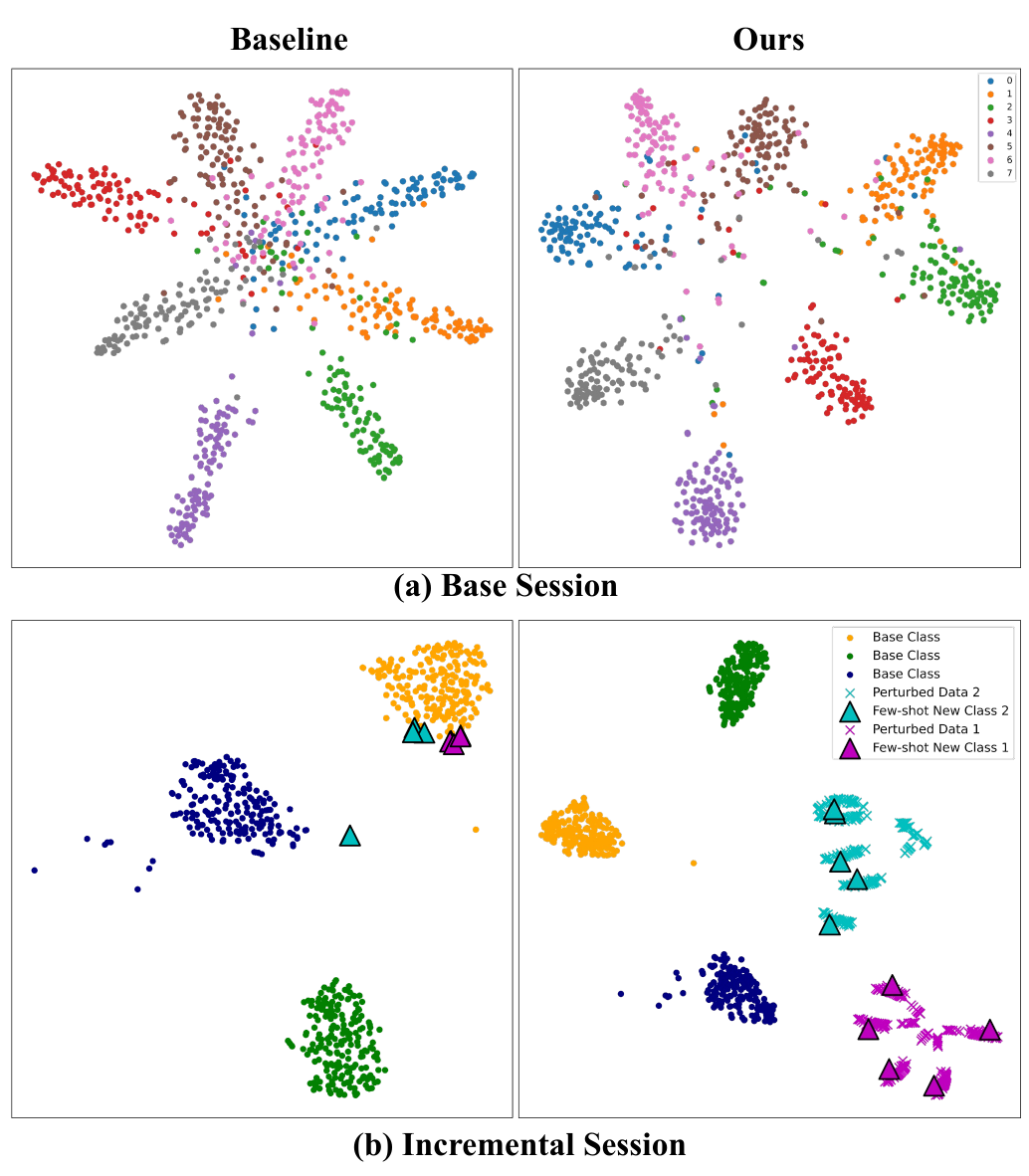}
    \caption{Comparisons of T-SNE visualization of the learned feature embedding on CIFAR100. (a) Visualization of eight base classes' feature embedding in the base session.  We compare the base model with and without covariance constraint loss. (b) Visualizations of the feature embedding of two few-shot new classes together with their perturbed samples, and three base classes during incremental learning. We compare the baseline model and the model using the proposed SPL. }
    \label{fig_tsne}
\end{figure}

We compare the different approaches by visualizing the learned feature embedding on the CIFAR100 dataset. In \cref{fig_tsne}(a), we compare the base model trained with and without covariance constraint loss. We visualize the feature embedding of 8 different base classes, demonstrating that by adding the constraint on the covariance of the feature distribution, the learned feature distribution becomes more separable and compact, leaving more feature space for the incoming new classes. In \cref{fig_tsne}(b), we demonstrate the effectiveness of SPL by showing the learned new class features before SPL,  and new class features along with all perturbed data after SPL during incremental sessions $(t>0)$. As shown in the left side of \cref{fig_tsne}(b), the new class features (colored triangles) tend to lie close with each other and mix with base classes due to the scarcity of the training data. By deploying SPL during the incremental sessions, our approach is able to push the few-shot training samples away from base classes and facilitate better separation among classes. The generated perturbed data in each epoch (denoted by ``$\times$" in the figure) expands the original small new class feature distribution, which allows the new class distributions to take more feature space, benefiting the new class prediction.

\subsection{Further Analysis}

\subsubsection{Effects of Incremental Shots}
We further conduct experiments on CUB200 to investigate the impact of varying the number of shots during incremental sessions, as shown in \cref{fig_further_anaylsis}(a). Since our method relies on the data of each class to expand the new class distributions, we vary the number of shots per class to observe its effect in incremental sessions. Keeping the number of classes consistent with the current experimental setting, we vary the shot number $K$ from $\{1, 3, 5, 10\}$ on CUB200. As depicted in the figure, increasing the number of available instances in each class leads to more accurate distributions and a corresponding improvement on performance. Even with a reduced shot number of 1, the model maintains stable performance during incremental sessions.

\subsubsection{Hyperparameter Analysis}
\label{abl_hyper}
\begin{figure}[ht]
    \centering
    \includegraphics[width=\linewidth]{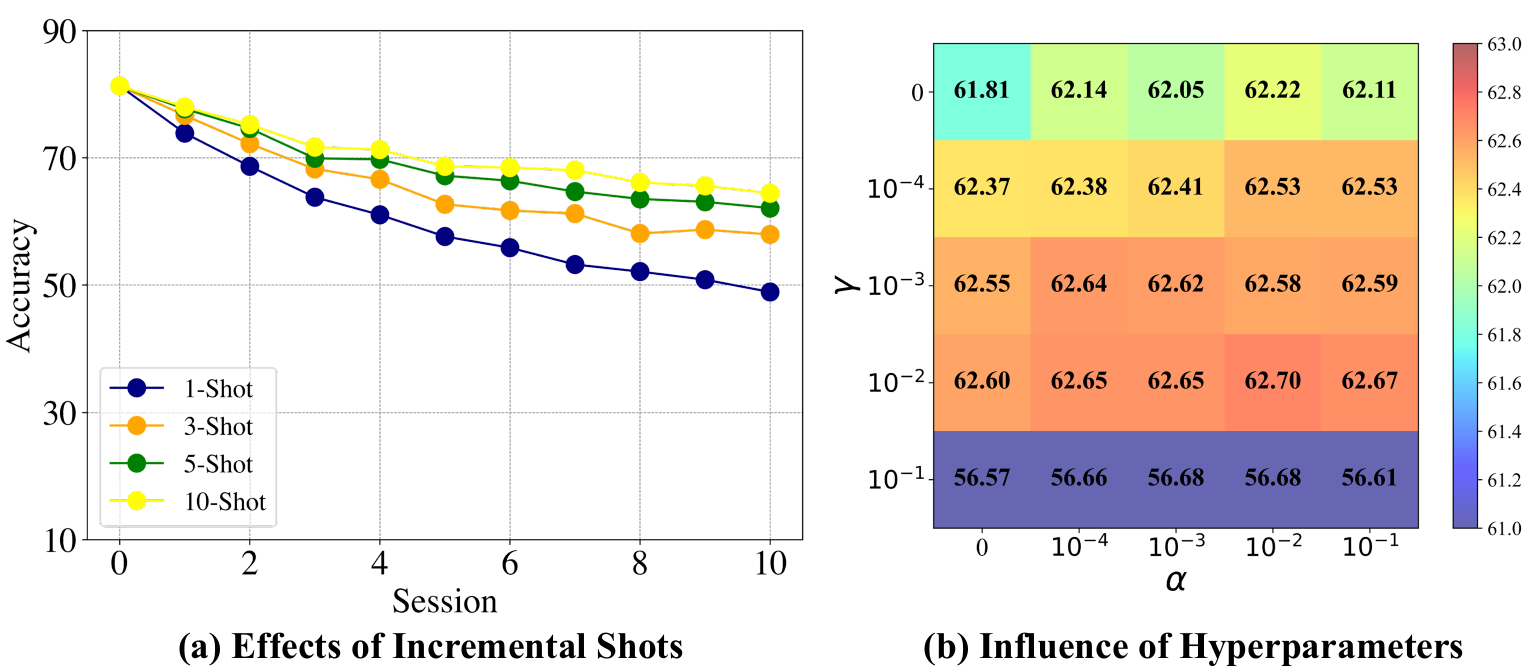}
    \caption{(a) Results of different incremental shots on CUB200. (b) Results of different hyperparameters on CUB200.}
    \label{fig_further_anaylsis}
\end{figure}

There are two hyperparameters in our approach, i.e., the $\gamma$ controls the impact of covariance constraint loss in \cref{eq_base_loss} and the $\alpha$ controls the perturbation in \cref{eq_spl}.  We conduct experiments by changing $\gamma$ and $\alpha$ from $\{0,0.0001,0.001,0.01,0.1\}$, and show the accuracy of the last session on CUB200 dataset in \cref{fig_further_anaylsis} (b). We can see that relatively small $\gamma$ and $\alpha$ yield better performance, and the best performance is yield with $\gamma$ and $\alpha$ set as 0.01. 

\section{Conclusion}
In this paper, we propose a covariance constraint loss and semantic perturbation learning to address the ill-divided feature space problem for few-shot class incremental learning (FSCIL). Our main motivation is to constrain the span of each distribution then reallocate the ill-divided feature space by 
reestablishing the decision boundaries between classes. Based on this, we attempt to learn the model through two steps: during the base session, we propose a covariant constraint loss (CCL) to explicitly constrain the feature distribution and facilitate better class separation. We derive the CCL from a variational inference framework, which estimates and constrains the feature distribution efficiently. For incremental sessions, we propose to generate semantic-guided perturbed data to aid the learning of few-shot new classes. The generated data expands the few-shot distributions and pushes the few-shot samples away from easily confusing classes.  The proposed approach can be integrated into current FSCIL methods in a plug-and-play manner, which is easy to implement. We conduct comprehensive experiments on three benchmark datasets and apply our approach to three baseline models. Experimental results demonstrate the effectiveness of our methods and obtain a new state-of-the-art performance.

\section{Acknowledgements}
The work was partially supported by the following: National Natural Science Foundation of China under no. 92370119 and No. 62276258, and No.62376113.

{\small
\bibliographystyle{ieee_fullname}
\bibliography{icdm_reference}
}

\appendix
\renewcommand*{\thefigure}{A\arabic{figure}}
\renewcommand*{\thetable}{A\arabic{table}}

\newpage
\twocolumn



This Appendix first provides details of the three baseline models in \cref{Details_of_Baseline_Models}. Comparisons of different incremental learning approaches with SPL are provided in \cref{abl_incremental}. In \cref{more_experiments}, we provide  additional experimental results on CIFAR100 and
CUB200.
\section{Details of Baseline Models}
\label{Details_of_Baseline_Models}
We choose three prevalent baseline models, i.e., the naive CE-trained model, fantasy-based model FACT~\cite{zhou2022forward}, and SAVC as the baseline models~\cite{song2023learning}.
\begin{itemize}
    \item \textbf{CE}: The base model is trained simply using cross-entropy loss in the base session. For the incremental sessions, the feature extractor is frozen, and only the classifier is updated.

    \item \textbf{FACT}~\cite{zhou2022forward}: During the base session, virtual new classes are synthesized by manifold mixup~\cite{verma2019manifold} to assist the base training, intending to save feature space for new classes. For incremental sessions, the model is updated by adding new prototypes to the classifier.

    \item \textbf{SAVC}~\cite{song2023learning}: Contrastive learning~\cite{khosla2020supervised} is adopted in base session to learn compact representations.  During the incremental sessions, multiple prototypes from each new class are ensembled as new classifier parameters.
\end{itemize}

\section{Comparison of Different Incremental Learning Approaches}
\label{abl_incremental}
We conduct experiments on comparing different incremental learning approaches on fine-grained dataset CUB200 to verify the effectiveness when learning new classes. As shown in \cref{tab_incremental}, we compare the proposed SPL with two commonly used incremental learning approaches, i.e., prototype-based update in \cref{Prototype-based Model}, fine-tuning the last layer of the model by CE using few-shot data~\cite{yang2023neural,song2023learning}. We use the same base model, followed by 10 incremental sessions.
The prototype-based model obtains the lowest new class performance, average performance of ten sessions, and harmonic accuracy since it cannot separate the new classes from other classes efficiently. The prototype-based model does obtain the highest old class performance and the least drop in PD, as it does not involve any update of feature space. The finetune approach boosts the new class performance by updating the feature space, hence obtaining higher average performance and harmonic accuracy. Compared to the finetuning approach, our SPL expands the new class feature distributions and facilitates a wider margin between classes. Therefore, SPL can retrain higher base class performance while learning new classes more effectively, achieving the highest performance of new classes, average performance and harmonic accuracy.

\begin{table}[ht]
\caption{Comparison of different incremental Methods. ``Prototype-based'' refers to the approach that simply updates the new class prototypes during incremental learning.   ``Finetune by CE'' denotes using CE to finetune the last layer of the model with few-shot data. }
\resizebox{\linewidth}{!}{
\begin{tabular}{@{}c|cccccc@{}}
\toprule
\multirow{2}{*}{\textbf{Incremental Method}} & \multicolumn{6}{c}{\textbf{CUB200}} \\ \cmidrule(l){2-7} 
 & \textbf{Base$\uparrow$} & \textbf{Old$\uparrow$} & \textbf{New$\uparrow$} & \textbf{Avg$\uparrow$} & \textbf{PD$\downarrow$} & \textbf{H. $\uparrow$} \\ \midrule
Prototype-based~\cite{zhou2022forward} & 81.31 & \textbf{76.96} & 47.00 & 68.88 & \textbf{4.35} & 58.35 \\
Finetune by CE~\cite{song2023learning} & 81.31 & 76.54 & 47.71 & 68.93 & 4.77 & 58.78 \\
\rowcolor{gray!20}
SPL & 81.31 & 76.68 & \textbf{47.85} &\textbf{69.12} & 4.63 & \textbf{58.93} \\ \bottomrule
\end{tabular}}
\label{tab_incremental}
\end{table}

\section{More Benchmark Results}
\label{more_experiments}

\begin{table*}[ht]
\centering
\caption{
Incremental learning performance on CIFAR100 under 5-way 5-shot setup. “Avg Acc.” represents the average accuracy of all sessions. “Final Improv.” calculates the improvement of our method after learning in the final session. \textbf{Bold} represents best performance. $*$ indicates that we reproduce the results using public open-source code}
\resizebox{\linewidth}{!}{
\begin{tabular}{@{}clccccccccccc@{}}
\toprule
\multicolumn{2}{c}{\multirow{2}{*}{\textbf{Methods}}} & \multicolumn{9}{c}{\textbf{Accuracy in each session (\%) $\uparrow$}} &  \multirow{2}{*}{\textbf{\begin{tabular}[c]{@{}c@{}}Avg\\  Acc.\end{tabular}}} & \multirow{2}{*}{\textbf{\begin{tabular}[c]{@{}c@{}}Final\\ Improv.\end{tabular}}}  \\ \cmidrule(lr){3-11}
\multicolumn{2}{c}{} & \textbf{0} & \textbf{1} & \textbf{2} & \textbf{3} & \textbf{4} & \textbf{5} & \textbf{6} & \textbf{7} & \textbf{8} &  &  \\ \midrule
 & iCaRL~\cite{rebuffi2017icarl} & 64.10 & 53.28 & 41.69 & 34.13 & 27.93 & 25.06 & 20.41 & 15.48 & 13.73 & 32.87 & +39.41 \\
 & NCM~\cite{hou2019learning} & 64.10 & 53.05 & 43.96 & 36.97 & 31.61 & 26.73 & 21.23 & 16.78 & 13.54 & 34.22 & +39.60 \\ 
 & Data-free Replay~\cite{liu2022few} & 74.40 & 70.20 & 66.54 & 62.51 & 59.71 & 56.58 & 54.52 & 52.39 & 50.14 & 60.78 & +3.00\\
 & Self-promoted~\cite{zhu2021self} & 64.10 & 65.86 & 61.36 & 57.45 & 53.69 & 50.75 & 48.58 & 45.66 & 43.25 & 54.52 & +9.89 \\
 & CEC~\cite{lesort2020continual} & 73.07 & 68.88 & 65.26 & 61.19 & 58.09 & 55.57 & 53.22 & 51.34 & 49.14 & 59.53 & +4.00 \\
 & MetaFSCIL~\cite{chi2022metafscil} & 74.50 & 70.10 & 66.84 & 62.77 & 59.48 & 56.52 & 54.36 & 52.56 & 49.97 & 60.79 & +8.05 \\
 & C-FSCIL~\cite{hersche2022constrained} & 77.47 & 72.40 & 67.47 & 63.25 & 59.84 & 56.95 & 54.42 & 52.47 & 50.47 & 61.64 & +3.17 \\
&LIMIT~\cite{zhou2022few} & 72.32 & 68.47 & 64.30 & 60.78 & 57.95 & 55.07 & 52.70 & 50.72 & 49.19 & 59.06 & +3.95 \\ \midrule
& CE  & 76.87 & 72.38 & 68.06 & 63.83 & 60.52 & 57.76 & 55.47 & 53.25 & 50.94&62.12&+2.20\\
\rowcolor{gray!20}
&\textbf{CE-Ours} & 78.27 & 73.80 & 69.69 & 65.53 & 62.07 & 59.33 & 57.22 & 54.75 & 52.30 & 62.21&+0.84\\\midrule
& FACT$^{*}$~\cite{zhou2022forward}& 78.38 & 71.86 & 67.87 & 64.10 & 60.70 & 57.75 & 55.83 & 53.6 & 51.34 & 62.17&+1.80\\
\rowcolor{gray!20}
& \textbf{FACT$^{*}$-Ours}& \textbf{79.12}& 72.62 & 68.49 & 64.31 & 61.51 & 58.64 & 56.38 & 54.22 & 52.34 & 62.82 &+0.80\\\midrule
&SAVC$^{*}$~\cite{song2023learning}& 78.98 & 73.02 & 68.69 & 64.49 & 60.91 & 58.08 & 55.79 & 53.61 & 51.75 & 62.81 &+1.39  \\
\rowcolor{gray!20}
& \textbf{SAVC$^{*}$-Ours} & 79.00 & \textbf{73.29} & \textbf{68.84} & \textbf{64.75} & \textbf{61.60} & \textbf{58.74} & \textbf{56.84} & \textbf{55.12} & \textbf{53.14} & \textbf{63.48}&\\
 \bottomrule
\end{tabular}
}
\label{tab_cifar}
\end{table*}

\begin{table*}[ht]
\caption{Performance of FSCIL in each session on CUB200 under 10-way 5-shot setup and comparison with other studies. “Average Acc.” is the average accuracy of all sessions. “Final Improv.” calculates the improvement of our method in the last session. \textbf{Bold} represents best performance. $*$ indicates that we reproduce the results using public open-source code.
}
\centering
\resizebox{\linewidth}{!}{
\begin{tabular}{@{}llccccccccccccc@{}}
\toprule
\multicolumn{2}{c}{\multirow{2}{*}{\textbf{Methods}}} & \multicolumn{11}{c}{\textbf{Accuracy in each session (\%) $\uparrow$}} &  \multirow{2}{*}{\textbf{\begin{tabular}[c]{@{}c@{}}Avg\\  Acc.\end{tabular}}} & \multirow{2}{*}{\textbf{\begin{tabular}[c]{@{}c@{}}Final\\ Improv.\end{tabular}}}  \\ \cmidrule(lr){3-13}
\multicolumn{2}{c}{} & \textbf{0} & \textbf{1} & \textbf{2} & \textbf{3} & \textbf{4} & \textbf{5} & \textbf{6} & \textbf{7} & \textbf{8}&\textbf{9}&\textbf{10} &  &  \\ \midrule
&iCaRL~\cite{rebuffi2017icarl} & 68.68 & 52.65 & 48.61 & 44.16 & 36.62 & 29.52 & 27.83 & 26.26 & 24.01 & 23.89 & 21.16 & 36.67 &+41.54\\
&Data-free Replay~\cite{liu2022few} & 75.90 & 72.14 & 68.64 & 63.76 & 62.58 & 59.11 & 57.82 & 55.89 & 54.92 & 53.58 & 52.39 & 61.52 &+10.31\\
&LDC~\cite{liu2023learnable}&77.89 &76.93 &74.64& 70.06& 68.88 &67.15 &64.83& 64.16 &63.03& 62.39 &61.58&68.32&+1.12\\
& CEC~\cite{lesort2020continual}& 75.85 & 71.94 & 68.50 & 63.50 & 62.43 & 58.27 & 57.73 & 55.81 & 54.83 & 53.52 & 52.28 & 61.33&+10.42\\
& LIMIT~\cite{zhou2022few} & 76.32 & 74.18 & 72.68 & 69.19 & 68.79 & 65.64 & 63.57 & 62.69 & 61.47 & 60.44 & 58.45 & 66.67&+4.25\\
& MetaFSCIL~\cite{chi2022metafscil} & 75.90 & 72.41 & 68.78 & 64.78 & 62.96 & 59.99 & 58.3 & 56.85 & 54.78 & 53.82 & 52.64 & 61.93&+10.06 \\ \midrule
& CE & 79.32 & 75.67 & 72.56 & 67.42 & 66.46 & 62.00 & 60.85 & 59.31 & 57.78 & 56.88 & 55.73 & 64.91&+6.97\\
\rowcolor{gray!20}& \textbf{CE-Ours} & 79.59 & 75.32 & 72.31 & 67.46 & 66.68 & 63.61 & 62.68 & 61.07 & 59.09 & 59.20 & 58.34  & 65.71&+4.36\\ \midrule
&  FACT$^{*}$ &77.28 & 73.67 & 70.19 & 65.59 & 64.77 & 61.60 & 60.68 & 58.89 & 57.38 & 57.26 & 56.11 & 63.87&+6.59\\
\rowcolor{gray!20}& 
\textbf{FACT$^{*}$-Ours}&77.78 & 74.23 & 70.42 & 65.97 & 65.31 & 61.58 & 61.42 & 59.61 & 57.42 & 57.26 & 56.49 & 65.15&+6.21 \\  \midrule
&SAVC$^{*}$ & 81.31 &77.35 & 74.49& 69.65 & 69.78 & 67.10 & 66.48 & 64.09 & 63.16 & 62.48 & 61.81 & 68.88&+0.89\\
\rowcolor{gray!20}&
\textbf{SAVC$^{*}$-Ours} & \textbf{82.67} & \textbf{78.58} & \textbf{75.66} & \textbf{70.83} & \textbf{70.37} &\textbf{67.30} & \textbf{66.80} & \textbf{65.57} & \textbf{64.01} & \textbf{63.45} & \textbf{62.70} & \textbf{69.81} \\
 \bottomrule
\end{tabular}
}
\label{tab_cub}
\end{table*}

We also present the performance of our method on the CIFAR100 and CUB200 datasets, as shown in \cref{tab_cifar} and \cref{tab_cub}, respectively. On CIFAR100, our approach  boosts the performance of baseline methods in all sessions. Our method improves the final performance of three baselines by at least 0.65\% and boosts the average performance on all incremental sessions. The improvement is attributed to the covariance constraint loss and semantic perturbation learning, which promote effective class separation and few-shot new class learning. On the fine-grained dataset CUB200, which includes 200 classes, our method achieves a final performance of 62.70\%, demonstrating the effectiveness of our approaches. We obtain an improvement in final accuracy of 2.61\% by applying our approach to the CE baseline model. In session 1 and session 2, our method yields lower performance on the CE model due to the imbalance of base class and new class in the testing data, but in the following incremental sessions, our approach is able to boost the overall performance.

\end{document}